\documentclass[10pt,twocolumn,letterpaper]{article}

\usepackage{iccv}
\usepackage{times}
\usepackage{epsfig}
\usepackage{graphicx}
\usepackage{amsmath}
\usepackage{amssymb}

\usepackage{booktabs}
\usepackage{multirow}
\usepackage{makecell}
\usepackage{float}

\usepackage{capt-of}
\newcommand*{\x}{\mathsf{x}\mskip1mu}

\usepackage[pagebackref=true,breaklinks=true,letterpaper=true,colorlinks,bookmarks=false]{hyperref}

\usepackage[capitalize]{cleveref}
\crefname{section}{Sec.}{Secs.}
\Crefname{section}{Section}{Sections}
\Crefname{table}{Table}{Tables}
\crefname{table}{Tab.}{Tabs.}

\iccvfinalcopy % *** Uncomment this line for the final submission

\def\iccvPaperID{5938} % *** Enter the ICCV Paper ID here

\begin{document}

%%%%%%%%% TITLE
\title{Diffusion-SDF: Conditional Generative Modeling of Signed Distance Functions}

\author{Gene Chou\\
Princeton University\\
{\tt\small gchou@princeton.edu}
\and
Yuval Bahat\\
Princeton University\\
{\tt\small yb6751@princeton.edu}
\and
Felix Heide\\
Princeton University\\
{\tt\small fheide@princeton.edu}
}

\maketitle
\ificcvfinal\thispagestyle{empty}\fi

\begin{figure*}[t]
    \centering 
    \includegraphics[width=\linewidth]{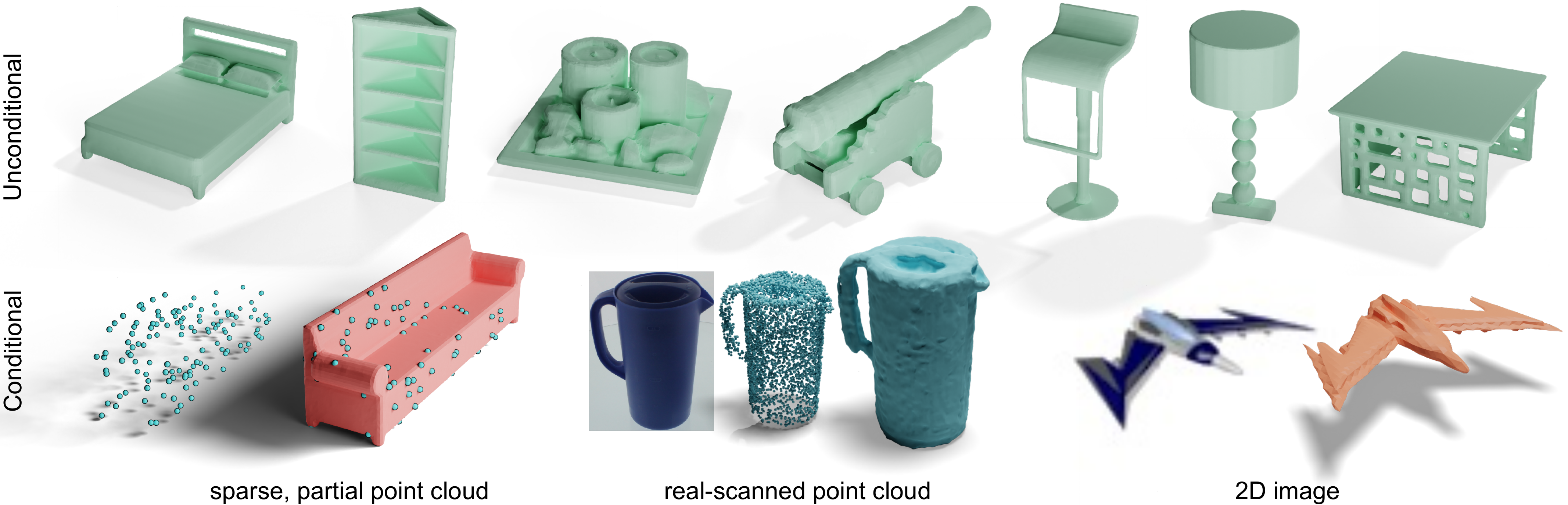}
    \caption{Our method generates clean meshes with diverse geometries. (\textbf{Top}) Unconditional generations from training on multiple classes. (\textbf{Bottom}) Conditional generation given various visual inputs, such as partial point clouds (same point cloud overlaid on sample), real-scanned point clouds, and 2D images. Our method captures details of conditioned geometry, such as the handle of the pitcher.}
    \label{fig:teaser}
\end{figure*}

%%%%%%%%% ABSTRACT
\begin{abstract}
    Probabilistic diffusion models have achieved state-of-the-art results for image synthesis, inpainting, and text-to-image tasks. However, they are still in the early stages of generating complex 3D shapes. 
    This work proposes Diffusion-SDF, a generative model for shape completion, single-view reconstruction, and reconstruction of real-scanned point clouds. 
    We use neural signed distance functions (SDFs) as our 3D representation to parameterize the geometry of various signals (e.g., point clouds, 2D images) through neural networks.
    Neural SDFs are implicit functions and diffusing them amounts to learning the reversal of their neural network weights, which we solve using a custom modulation module. 
    Extensive experiments show that our method is capable of both realistic unconditional generation and conditional generation from partial inputs.
    This work expands the domain of diffusion models from learning 2D, explicit representations, to 3D, implicit representations. 
\end{abstract}

%%%%%%%%% BODY TEXT
\vspace{-3em}
\section{Introduction}
\label{sec:intro}

Diffusion probabilistic models~\cite{sohl2015deep, ddpm} have become a popular choice for generative tasks and can produce impressive results, such as the images generated by DALLE-2~\cite{dalle2} and Stable Diffusion~\cite{stable-diffusion} from text input. Diffusion models are a type of likelihood-based models whose training objective can be expressed as a variational lower bound \cite{ddpm, score_functions}. On a high level, they learn to gradually remove noise from a signal and repeat this process to generate samples from Gaussian noise. Recent advances~\cite{improved_ddpm, diffusion_beat_gans, dalle2, stable-diffusion} show that diffusion models produce images with quality on par with state-of-the-art generative adversarial networks (GANs)~\cite{gan} without the common drawbacks of mode collapse~\cite{improved_ddpm, nash2021generating} and unstable training~\cite{spectralgan, largegan}. Diffusion has also been applied to 3D tasks although these works are still in the early stages of producing complex shapes. In this work, we investigate the generation of 3D shapes of neural signed distance functions via diffusion.

3D modeling and generation are essential to vision and graphics tasks. 3D generation of high-quality assets and large volumes of realistic data is often essential where training data is expensive to collect~\cite{CT,drug-design,facial-3d, facial-privacy, asset-games, asset-archaeology}. Additionally, generation can be applied to 3D reconstruction of imperfect visual observations as there exists a one-to-many mapping that requires a probabilistic approach to solve. This has applications in self-driving~\cite{waymo, star-tracking, neuralscenegraph} and robotics grasping~\cite{ycb, grasping-survey, nerfsupervision} where occlusion and camera measurement errors are common. 

We propose Diffusion-SDF, a generative model for shape completion, single-view reconstruction, and reconstruction of real-scanned point clouds. We choose neural signed distance functions (SDFs)~\cite{deepsdf} as the 3D representation to parameterize the surfaces described by various input signals such as point clouds and 2D images. They implicitly encode an object surface by the signed distances between 3D coordinate queries to their closest surface point through a coordinate-based MLP~\cite{deepsdf,gensdf}. Compared to discrete 3D representations~\cite{voxfields, voxnet, pointnet, pointcloud, atlasnet}, SDFs have proven to be a versatile representation that supports arbitrary resolution during test-time~\cite{nglod}, small memory footprints~\cite{singleshapesdf}, and strong generalization~\cite{gensdf}. 

We make the following two key insights. First, implicit functions can directly be used as data and diffusing them amounts to learning the reversal of the neural network weights. Furthermore, we introduce geometrical constraints to produce complex shapes and outputs consistent with the geometry of conditioned inputs. Very recently, Dupont \etal~\cite{functa} similarly diffuse implicit functions but do not address SDFs nor geometric constraints. Second, by using SDFs as a unified 3D representation, we condition training to learn a mapping between various input types and their possible reconstructions. Then, we leverage a probabilistic diffusion model to generate diverse completions. Thus, our work can be applied to shape synthesis and multi-modal shape completion.

\begin{figure*}[t]
    \centering 
    \includegraphics[width=0.9\linewidth]{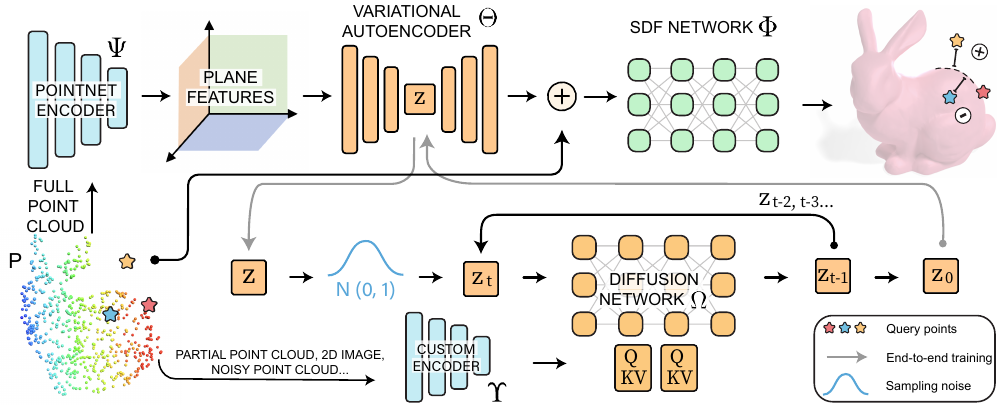}
    \caption{Our two-stage training pipeline. The first (\textbf{top}) trains SDFs jointly with a VAE~\cite{vae} to produce latent vectors $\textbf{z}$ each representing an SDF embedding. The second stage (\textbf{bottom}) uses the latent vectors as input to our diffusion model and can be guided by various inputs. We connect the two models (\textbf{gray arrow}) for end-to-end training. During test time, the diffusion model takes input $z$ sampled from a Gaussian distribution and we combine its output with the SDF network to form a complete SDF representation.}
    \label{fig:training}
\end{figure*}

The proposed method consists of two steps, shown in \cref{fig:training}. First, we create a compressed representation of SDFs using modulation~\cite{functa, pigan, mod-periodic}. We find that diffusing SDFs is impractical due to the large number of parameters and the lack of a smoothed data distribution. Our modulation module consists of learning a generalizable encoder and a regularized latent space to create latent vectors that map to individual SDFs when combined with an SDF base network. Second, we train a diffusion model with the previously created latent vectors as data points. We follow the conventional approach of learning the reverse diffusion process~\cite{ddpm, dalle2}, but we combine it with our modulation scheme to introduce geometric information. We show that this geometric constraint is essential for the method to complete shapes consistent with guided inputs. Furthermore, we experiment with various input types for guiding generation. Shown in \cref{fig:teaser}, we validate the method by shape generation and completion with conditioning of partial point clouds from Acronym~\cite{acronym}, real-scanned point clouds from YCB~\cite{ycb}, and 2D images from ShapeNet~\cite{shapenet}. Our method generates diverse and realistic shapes for multiple tasks. Code and supplement can be found at \url{https://light.princeton.edu/publication/diffusion-sdf/}.

We make the following contributions:
\begin{itemize}
    \vspace{-0.5em}
    \item We propose a probabilistic generative model that creates clean and diverse 3D meshes.
    \vspace{-0.5em}
    \item We solve a learning problem of diffusing the weights of implicit neural functions while providing geometric guidance through our modulation module. 
    \vspace{-0.5em}
    \item Our method reconstructs plausible outputs from various imperfect observations such as sparse, partial point clouds, single images, and real-scanned point clouds.
    \vspace{-0.5em}
    \item Extensive experiments show that our method achieves favorable performance in shape generation and completion compared to existing methods.
\end{itemize}

\section{Related Work}
\label{sec:related_work}

\paragraph{Diffusion Probabilistic Models}

Diffusion probabilistic models~\cite{sohl2015deep, ddpm} generate samples from a distribution by learning to gradually remove noise from a datapoint. 
% There are different variations, but the following describes a canonical one. We denote a sampled datapoint as $x_0$, and iteratively add to it a small Gaussian noise to obtain $x_1, x_2...x_T$, until $x_T$ approximates an isotropic Gaussian. This forward step is a Markovian fixed process~\cite{ddpm, score_functions}. Then, for generation we start with $x_T \sim \mathcal{N}(0,\textbf{I})$, and iteratively denoise $x_T$ with ancestral sampling. 
Recent advances~\cite{improved_ddpm, diffusion_beat_gans, dalle2, stable-diffusion} show diffusion models produce high quality images without the drawbacks of mode collapse~\cite{improved_ddpm, nash2021generating} and unstable training~\cite{spectralgan, largegan}. Diffusion has also been applied to 3D tasks although these works are still in the early stages of producing complex shapes. 

One line of work~\cite{pvd, pc-diffusion, lion} trained diffusion models to generate point clouds. Very recently, Dupont \etal~\cite{functa} trained diffusion models on implicit neural representations but not SDFs. These existing methods perform unconditional generation and some only produce simple geometries. Our modulation scheme and conditioning mechanisms fix both of these issues. We also acknowledge concurrent works that combine diffusion and implicit functions; \cite{dreamfusion, 3DiM} generate novel views from text and images with and without intermediate radiance fields, respectively. 
%These concurrent approaches do directly generate shape or operate on SDFs as the proposed method. %Hui et al.~\cite{wavelet} trains diffusion models for signed distance functions (SDFs), but their SDFs are first transformed into wavelet representations, which means the representation capacity is limited by the transformation. Our method directly learns the reversal of the SDF weights and is trained end-to-end to learn complex geometries.

\paragraph{Generative Modeling of 3D Shapes}
Many existing works~\cite{if-net, patchcomplete, pointr, cascaded-point-completion} that reconstruct partial scans and meshes are deterministic, but the relation between partial and completed shapes is a one-to-many mapping. To address this, \cite{autosdf, shapeformer, cgan, pvd} proposed probabilistic models for generating multi-modal reconstructions that are consistent with partial inputs. Mittal \etal~\cite{autosdf} train an autoregressive prior, Wu \etal~\cite{cgan} train a conditional GAN, Yan \etal~\cite{shapeformer} train a vector quantized deep implicit function (VQDIF), and Zhou \etal~\cite{pvd} train a diffusion model. We compare our method to these baselines. 

More recently, some works~\cite{get3d, wavelet} have combined SDFs and generative modeling. Gao \etal~\cite{get3d} train on 2D image collections and combine differentiable rendering and 2D GANs. Hui \etal~\cite{wavelet} is concurrent work that learns diffusion models. They convert SDFs into wavelet representations, then use them as input to diffusion models. Different from the proposed approach, this method cannot perform conditional generation and learns to reverse wavelets as a surrogate to implicit functions.

\paragraph{Learning Implicit Signed Distance Functions}
A rapidly growing body of work relies on implicit neural networks as an expressive scene representation that facilitates learning for 3D reconstruction and view synthesis tasks~\cite{deepsdf, occupancy, nerf}. They use neural networks to map spatial coordinates to scene attributes, which offers a fully-differentiable and versatile way to represent 3D geometry. Park \etal~\cite{deepsdf} first proposed to map coordinates to signed distance values and reconstructed surfaces by interpolating grid points with signed distance values of zero.

Follow-up works~\cite{gensdf, conv_occ, neuralpull} use point clouds as additional conditions to achieve greater detail, generalization, and unsupervised training. However, they condition their input on \emph{full-view} point clouds with low levels of noise. Thus, they fail on real-world applications where one can only obtain partial and noisy point clouds. In this work, we close this gap and introduce a generative model that reconstructs plausible outputs for partial and noisy point clouds. 
%real-world applications such as autonomous driving~\cite{waymo} and robotics manipulation~\cite{ycb}. 

% \paragraph{Modulation and Compression of Data}
% Training diffusion models can become computationally expensive when input data has high dimensionality~\cite{ddpm, stable-diffusion}. Rombach \etal~\cite{stable-diffusion} train an autoencoder to use a compressed representation of high-resolution images. Hui \etal~\cite{wavelet} transform SDFs into wavelet representations. Dupont~\etal~\cite{functa} map latent codes linearly to a base implicit function, allowing the diffusion model to take latent vectors, instead of entire neural networks, as input. 

% However, the learned mappings of Dupont~\etal and Hui~\etal only fit simple geometries and few categories. In contrast, the proposed method for modulating implicit functions combines autoencoders and a mapping between a latent code and a base network, and is capable of compressing and reconstructing hundreds of categories with complex geometries. Additionally, our modulation module fits into our training of SDFs which allows the flow of geometrical constraints between diffusing neural network weights and the generation of modulation vectors. 

\section{Diffusion Models and Neural SDFs}
\label{sec:background}

\paragraph{Background on Diffusion Probabilistic Models}
While different variations of diffusion models exist, we describe a canonical one~\cite{sohl2015deep, ddpm}. From a data distribution $q(\mathbf{x})$, we denote a sampled datapoint as ${x}_0$, and iteratively add small Gaussian noise to obtain ${x}_1, {x}_2...{x}_T$, until $x_T$ approximates an isotropic Gaussian. 
This forward step is a Markovian fixed process~\cite{ddpm, score_functions} and can be defined as $q({x}_{1: T} | {x}_0)=\prod_{t=1}^T q({x}_t | {x}_{t-1})$ and $q({x}_t | {x}_{t-1})=\mathcal{N}({x}_t ; \sqrt{1-\beta_t} {x}_{t-1}, \beta_t {I})$ where $\beta_t$ is a variance schedule. In practice, we sample ${x}_t$ using a closed form parameterization $\sqrt{\bar{\alpha}_t} {x}_0+\sqrt{1-\bar{\alpha}_t} {\epsilon}$ where $\alpha_t=1-\beta_t$, $\bar{\alpha}_t=\prod_{i=1}^t \alpha_i$, and ${\epsilon} \sim \mathcal{N}(0,{I})$.

The goal of each training iteration is to train a model $p_\theta$, often represented by a neural network, that inverts the forward diffusion (i.e., learns the \textit{reverse} diffusion process): $p_\theta({x}_{0: T})=p({x}_T) \prod_{t=1}^T p_\theta({x}_{t-1} | {x}_t)$ and $p_\theta({x}_{t-1} | {x}_t)=\mathcal{N}({x}_{t-1} ; {\mu}_\theta({x}_t, t), {\Sigma}_\theta({x}_t, t))$. The reverse process is also Markovian and we fix the variances ${\Sigma}_\theta$. The reverse conditional probability is tractable when conditioned on ${x}_0$: $q({x}_{t-1} | {x}_t, {x}_0)=\mathcal{N}({x}_{t-1} ; \tilde{{\mu}}({x}_t, {x}_0), \tilde{\beta}_t {I})$. We apply Bayes' rule to rearrange the terms and represent  $\tilde{{\mu}}({x}_t, {x}_0) = \frac{\sqrt{\alpha_t}(1-\bar{\alpha}_{t-1})}{1-\bar{\alpha}_t} {x}_t+\frac{\sqrt{\bar{\alpha}_{t-1}} \beta_t}{1-\bar{\alpha}_t} {x}_0$. The closed form parameterization of ${x}_t$ yields $\tilde{{\mu}}_t = \frac{1}{\sqrt{\alpha_t}}({x}_t-\frac{1-\alpha_t}{\sqrt{1-\bar{\alpha}_t}} {\epsilon}_t)$. 

Thus, we can train our model to predict $\tilde{{\mu}}_t$, or alternatively, ${\epsilon}_t$ by rearranging the terms. This work predicts $\tilde{{\mu}}_t$ for generating SDF samples.

For generation, we start with $x_T \sim \mathcal{N}(0,{I})$ and iteratively denoise $x_T$ with ancestral sampling~\cite{ddpm}: ${x}_{t-1}= {\mu}_\theta({x}_t, t) +\sigma_t {\epsilon}$ where $\sigma_t$ is the fixed standard deviation at timestep $t$ and  ${\epsilon} \sim \mathcal{N}(0,{I})$ is injected until the last step.

\paragraph{Background on Neural SDFs}
While many works overfit SDFs to a single object~\cite{singleshapesdf,nglod,siren}, some have been able to learn SDFs conditioned on point cloud inputs that generate shapes from different categories. A successful approach is jointly training a PointNet encoder~\cite{pointnet} and an SDF decoder~\cite{deepsdf, gensdf, conv_occ}, where shape features from the encoder are concatenated with 3D query points $x \in \mathbb{R}^3$ as input to the decoder. We denote the funcion $x, P \mapsto \Phi(x, P) = s$,
% \begin{equation}
%   x, P \mapsto \Phi(x, P) = s,
%   \label{eq:sdf-definition}  
% \end{equation}
where $P = \{p_i \in \mathbb{R}^3 \}_{i=1}^N$ is a raw point cloud with $N$ points, $\Phi : \mathbb{R}^3 \times \mathbb{R}^{3\times N}  \rightarrow \mathbb{R}$ is the SDF that predicts the signed distance value for a 3D coordinate, conditioned on a point cloud, and $s$ denotes the predicted signed distance value between $x$ and the shape described by $P$. The surface boundary of the shape is its zero-level set $S_0(\Phi(P))$, which can be formulated as $S_0(\Phi(P)) = \{z\in \mathbb{R}^3\ |\ \Phi(z, P)=0 \}$.
%\vspace{-1.5mm}
% \begin{equation}
%     S_0(\Phi(P)) = \{z\in \mathbb{R}^3\ |\ \Phi(z, P)=0 \} .
%     \label{eq:sdf-zero-level-set}
% \end{equation}
% Given a trained $\Phi$, we can visualize a surface by drawing its zero-level set. 

\section{Diffusing Neural Signed Distance Functions}
\label{sec:method}

% We first provide an overview and formulation of our method. 
% \paragraph{Overview} 
Illustrated in Fig.~\ref{fig:training}, the proposed method is composed of three major components: a modulation scheme to represent SDFs as individual latent vectors, a diffusion model that takes the latent vectors as distribution samples for training, and a custom encoder and attention mechanism for conditional generation. In the following, we describe the components of our method in detail.

\subsection{Modulating SDFs}
\label{sec:modulation}
We use modulation~\cite{functa, pigan, mod-periodic} to create an alternate representation of SDFs. Directly diffusing thousands of SDFs, where one SDF represents one object, is difficult because one must first train all SDFs separately (which would take thousands of GPU hours) and the distribution of thousands of SDFs is challenging to learn. We show in our supplement that existing diffusion models cannot directly learn from SDFs as training data. Thus, we map SDFs, represented by MLPs, to 1D latent vectors with two objectives: the diffusion model needs to learn and sample from the distribution of latent vectors effectively, and generated outputs of the diffusion model are mapped back into an SDF. This amounts to designing a latent space that needs to be continuous (interpolation between latent vectors corresponds to interpolation of geometry), complete (all points in the latent space are meaningful), and sufficiently diverse for holding information of hundreds of categories. 

To this end, we jointly train a conditional SDF representation and a VAE~\cite{vae}. We opt for the architecture of GenSDF~\cite{gensdf}, which is capable of learning hundreds of diverse categories using a unified model, so we train one model instead of thousands of SDFs. Specifically, our modulation module (see \cref{fig:training}), consisting of a PointNet encoder $\Psi$ and a VAE $\Theta$, takes in a raw point cloud $P = \{p_i \in \mathbb{R}^3 \}_{i=1}^N$ with $N$ points, and outputs plane features $\pi$ and $\pi^\prime$ and a latent vector $z$ as follows
\begin{equation}
  \pi = \Psi (P),\,\, z= \Theta_{enc} (\pi),\,\, \pi^\prime = \Theta_{dec} (z)
  \label{eq:modulation_output},
\end{equation}
where $\Theta_{enc}, \Theta_{dec}$ are the encoder and decoder of the VAE, respectively. Equivalently, $\pi^\prime = (\Theta \circ \Psi) (P)$. Other than compression, the VAE regularizes the latent space. Next, we pass the concatenation of query points $x \in \mathbb{R}^3$ and $\pi^\prime$ into the SDF network $\Phi$. We denote the predicted signed distance value as $s = \Phi(x|z)$. This formulation allows us to swap out different latent vectors $z$ for producing different shape representations, including generated latents from the diffusion model which we show later.

The training objective of this latent parameterization is to learn accurate predictions of signed distance values and regularize the latent space of the VAE as
\begin{equation}
  \mathcal{L}_{\mathrm{mod}} =  \| \Phi({x} | {z}) - \mathrm{SDF}({x})\|_1 + \beta(D_{KL}( q_\phi({z}|\pi) || p({z})))
  \label{eq:modulation}.
\end{equation}

The first term of the RHS of \cref{eq:modulation} is an $\mathcal{L}_1$ loss between the predicted and ground truth signed distance values of our query points ${x}$. Here, $\mathrm{SDF}(\cdot)$ denotes the ground-truth SDF operator that is defined for all $x \in \mathbb{R}^3$. The second term is our KL-divergence loss~\cite{vae} that regularizes the generated latent space to approach a target distribution. For given point cloud features $\pi$, we describe the inferred posterior of the latent vectors ${z}$ by a probability distribution $q_\phi({z}|\pi)$. We regularize the posterior to match a prior $p({z})$, which we set to be a Gaussian with zero-mean and standard deviation $0.25$. Diffusion processes converge toward Gaussian distributions so modeling data to approximate this distribution results in faster and more stable training. We also add a constant $\beta$ to control the strength of regularization, which we set to $1e{\text -}5$. We do not use a VAE reconstruction loss.

%supplement: Each shape feature has dimensions of $(3xDx64x64)$ -> explain how we concatenate the plane features 

Empirically our modulation method for implicit functions is capable of representing substantially more complex and diverse geometries compared to existing methods~\cite{functa, deepsdf}. We provide comparisons in the supplement.

\subsection{Diffusing Modulation Vectors}
Next, we use our sampled latent vectors ${z}$ from the previous step as sample space for the proposed diffusion probabilistic model $\Omega$, illustrated in \cref{fig:training}. In every iteration, Gaussian noise is added to the latent vectors at random timesteps, and the model learns to denoise the vectors. %We use the architecture of DALLE-2~\cite{dalle2}, with its open-source implementation from \cite{github-dalle2}. 
Instead of predicting the added noise $\epsilon$ as in the original DDPM~\cite{ddpm}, we follow Aditya \etal~\cite{dalle2} and predict $z_0$, the original, denoised vector. In other words, after we sample a timestep $t$ and noise $\epsilon$ to obtain $z_t$ from input latent vector $z_0$, the model learns to reconstruct $z_0$. 
The loss function is
\begin{equation}
  \mathcal{L}_{\mathrm{diff}} =  \| \Omega( z_t, \gamma( t)) -  z_0 \|_2 ,
  \label{eq:diffusion}
\end{equation}
where $\gamma (\cdot)$ is a positional embedding and $\| \cdot\|_2 $ is MSE loss.

We concatenate $ z_t$ and  $\gamma ( t)$ as input into the model, which has layers each consisting of attention~\cite{attention}, a fully connected layer, and normalization. We use the architecture of DALLE-2~\cite{dalle2} because they use 1D vectors as input, similar to our case. In contrast, the standard DDPM~\cite{ddpm} architecture is a UNet designed for images. During test time, our diffusion model performs generation iteratively as
\begin{equation}
  z^\prime = (f \circ ... \circ f)( z_T,  T), \,\,
  f(x_t, t) = \Omega( x_t, \gamma( t)) + \sigma_t  \epsilon,
  \label{eq:diffusion_generation}
\end{equation}
where $z_T \sim \mathcal{N}(0,{I})$, $\sigma_t$ is the fixed standard deviation at the given timestep, and $\epsilon \sim \mathcal{N}(0,{I})$. We iteratively denoise $z_T$ until we obtain the final output $z^\prime$.
Then, we pass the generated latent vectors $ z^\prime$ back into the joint SDF-VAE model for marching cubes reconstruction. 

\subsection{Conditioning Mechanisms}
% ADD CONDITIONAL DROPOUT!!!!!!!!!!!!!!!
% also remember to change generation process
One advantage of SDFs is their ability to represent 3D geometries from different modalities, such as point clouds and images~\cite{gensdf, imagetosdf-pretrain-selftrain}. Given some input $y$, we can train a custom encoder $\Upsilon$ to extract shape features $\pi = \Upsilon(y)$ to guide training of the diffusion model. We primarily experiment with partial point clouds, but in \cref{sec:modalities}, we show conditioning on real-scanned point clouds and 2D images. 

We use the same architecture for our diffusion model described in the previous subsection, but add $\Upsilon$ and an additional cross-attention layer to each block. 
Our cross-attention layer is the same as that used in \cite{stable-diffusion} and is defined as 
\begin{equation}
  \operatorname{Attention}(Q,K,V) = \operatorname{softmax}(\frac{QK^T}{\sqrt{d_k}})V
  \label{eq:attention}
\end{equation}
where $Q=W_Q^{(i)}\cdot \Omega_i(z_t, \gamma(t)), K=W_K^{(i)}\cdot \pi, V=W_V^{(i)}\cdot \pi$. $\Omega_i(\cdot)$ is the output of an intermediate layer of $\Omega$ and $W_Q, W_K, W_V$ are learnable matrices.
The cross-attention mechanism learns the mapping between the conditioned input and the geometry implicitly represented by the latent code. \cref{eq:diffusion} is now conditioned on $\pi$ and we have 
\begin{equation}
  \mathcal{L}_{\mathrm{c{\text -}diff}} =  \| \Omega( z_t, \gamma( t) |  \pi) -  z_0 \|_2 .
  \label{eq:cond_diffusion}
\end{equation}

Generation steps are the same as in \cref{eq:diffusion_generation} but in each step $\pi$ is given as a condition
\begin{equation}
  z^\prime = (g \circ ... \circ g)( z_T,  T,  \pi), \,\,
  g(x_t, t, \pi) = \Omega( x_t, \gamma( t) |  \pi) + \sigma_t  \epsilon
  \label{eq:diffusion_generation_conditioned}
\end{equation}
By conditioning our diffusion model during training, we can guide reconstructions during test time. Furthermore, we can generate \textit{multi-modal} reconstructions due to the generative nature of diffusion models. Finally, to increase diversity and prevent overfitting, we follow \cite{classifier-free-conditional-dropout}; every training iteration, with a certain probability we use a zero-mask instead of the shape feature as condition. In practice, we use the zero-mask with probability $80\%$.

\subsection{End-to-End Training for Geometry Constraints}
\label{sec:endtoend}
Our model consists of the creation of latent vectors through jointly training a conditional SDF and a VAE, and training the diffusion model using the latent vectors as input. These two modules can be trained end-to-end. As shown by the gray arrow in \cref{fig:training}, the output of the VAE can directly be used as input to the diffusion model, whose output can then be fed into the VAE decoder for calculating its SDF loss.
In practice, we found that training end-to-end from scratch took longer than training the modules separately since there are many building blocks. After the two modules complete training, however, we fine-tune them end-to-end. 
During training of the diffusion model individually, it can overfit to the input latent vectors since they are saved and fixed. When training end-to-end, the inputs are from the output of $(\Theta_{enc} \circ \Psi) (P)$ instead, which slightly vary each iteration, increasing generalization capabilities. Furthermore, the loss is originally based solely on the diffusion loss of the latent vectors, which does not have explicit geometrical constraints. By connecting the two modules, we introduce another SDF loss for the denoised latent vector, which guides SDF information to the diffusion model. This allows the model to generate more complex geometries. During this final stage of fine-tuning, we continue to use all the loss functions used in the separate modules, and add this additional constraint for end-to-end optimization 
\begin{equation}
  \mathcal{L}_{\mathrm{total}} = \mathcal{L}_{\mathrm{mod}} + \mathcal{L}_{\mathrm{c{\text -}diff}} + \| \Phi({x} |  z^\prime) - \mathrm{SDF}({x})\|_1,
  \label{eq:end-to-end}
\end{equation}
where $ z^\prime=\Omega( z_t, \gamma( t) |  \pi)$. We did not find it necessary to add weighing constants to the loss terms.

\begin{figure*}[t]
    \centering 
    \includegraphics[width=0.95\linewidth]{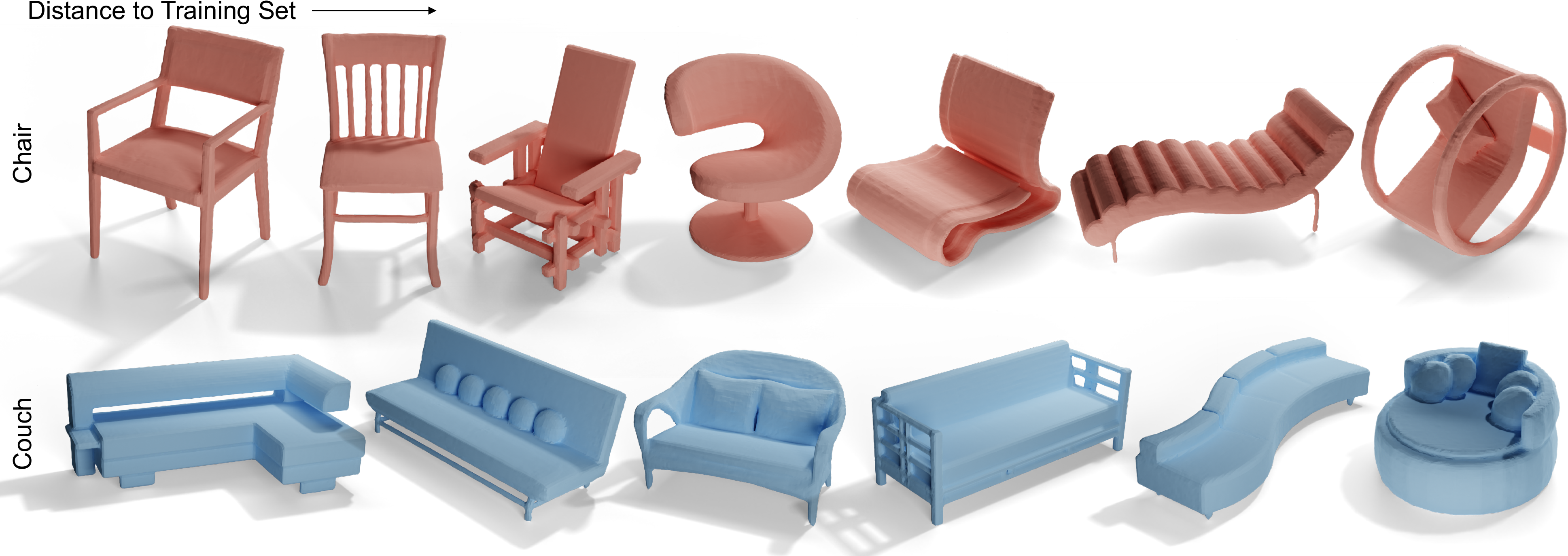}
    \caption{Samples from unconditional generation. Our method produces clean meshes with thin structures and diverse geometries. We also calculate their average CD to each object in the training set to confirm that our model is capable of producing unique shapes.}
    \label{fig:uncond-results}
\end{figure*}

\begin{table*}[ht!]
    \begin{minipage}{.5\linewidth}
    \centering
    \parbox{0.95\textwidth}{ \captionof{table}{Metrics for unconditional shape generation. $\uparrow$ means higher is better and $\downarrow$ means lower is better. MMD is scaled up by $10^2$. COV and 1-NNA are measured in percentages $(\%)$.  \label{tab:unconditional}}}
    \resizebox{0.95\linewidth}{!} 
     {%
     %\begin{table}
      \begin{tabular}{llccc}
        \toprule
        Shape & Model & MMD $(\downarrow)$ & COV $(\uparrow)$ & 1-NNA $(\downarrow)$ \\
        \midrule \multirow{4}{*}{Chair} 
        & ShapeGAN~\cite{shapegan} & $7.738$ & $8.661$ & $99.80$   \\
        & PVD~\cite{pvd} & $0.342$  & $39.43$  & $f86.56$  \\
        & DPM3D~\cite{pc-diffusion} & $0.130$ & $56.69$ &  $53.54$  \\
        & Ours & $\textbf{0.129}$ & $\textbf{65.35}$ &  $\textbf{51.18}$   \\
        \midrule \multirow{4}{*}{Couch} 
        & ShapeGAN~\cite{shapegan} & $6.527$ & $1.923$ &  $99.84$   \\
        & PVD~\cite{pvd} & $0.145$  & $49.45$ & $56.83$  \\
        & DPM3D~\cite{pc-diffusion} & $0.108$  & $48.72$ & $62.82$  \\
        & Ours & $\textbf{0.106}$  & $\textbf{61.22}$ & $\textbf{54.97}$   \\
       \midrule \multirow{4}{*}{\makecell{Multi-\\Class}} 
        & ShapeGAN~\cite{shapegan} & $4.659$ &$5.280$ & $99.99$   \\
        & PVD~\cite{pvd} & $0.350$ & $12.36$ & $93.33$\\
        & DPM3D~\cite{pc-diffusion} & $0.150$ & $45.40$ &  $68.36$    \\
        & Ours & $\textbf{0.131}$ & $\textbf{57.06}$ & $\textbf{67.38}$  \\
        \bottomrule
       
        \end{tabular}%
        }
    %\end{table}
    \end{minipage}%
    \begin{minipage}{.5\linewidth}
    \centering
    \parbox{0.95\textwidth}{\captionof{table}{Metrics for multi-modal shape completion of sparse, partial point clouds (128 points, $50\%$ cropped). $\uparrow$ means higher is better and $\downarrow$ means lower is better. All values are scaled up $10^2$. \label{tab:conditional}}}
      \resizebox{0.89\linewidth}{!} 
     {%
       \begin{tabular}{llccc}
        \toprule
        Shape & Model & MMD $(\downarrow)$ & TMD $(\uparrow)$ & UHD $(\downarrow)$ \\
        \midrule \multirow{4}{*}{Chair} & cGAN~\cite{cgan} & $0.193$ & $2.663$ & $7.804$  \\
        & PVD~\cite{pvd} & $0.504$ & $9.163$ & $\textbf{3.917}$  \\
        & SFormer~\cite{shapeformer} & $0.278$ & $4.820$ & $17.76$  \\
        & Ours &  $\textbf{0.036}$ & $\textbf{14.22}$ & $12.56$  \\
        \midrule \multirow{4}{*}{Couch} & cGAN ~\cite{cgan}&  $0.145$ & $2.231$ & $7.251$  \\
        & PVD~\cite{pvd} & $0.350$ & $7.920$ & $\textbf{6.134}$  \\
        & SFormer~\cite{shapeformer} & $0.103$ & $1.567$ & $7.270$  \\
        & Ours &  $\textbf{0.041}$ & $\textbf{13.53}$ & $10.37$  \\
       \midrule \multirow{4}{*}{\makecell{Multi-\\Class}} &cGAN~\cite{cgan} & $0.225$ & $1.994$ & $\textbf{7.162}$  \\
       & PVD~\cite{pvd} & $0.412$ & $10.16$ & $8.368$  \\
        & SFormer~\cite{shapeformer} & $0.208$ & $9.523$ & $14.98$  \\
        & Ours &  $\textbf{0.035}$ & $\textbf{20.11}$ & $14.86$  \\
        \bottomrule
        
        \end{tabular}%
    }
    \end{minipage} 
    \vspace{-1em}
\end{table*}

\section{Experiments}

Next, we validate the proposed method for generating shapes.  In \cref{sec:unconditional}, we report results of unconditional generation initialized from Gaussian noise. In \cref{sec:conditional} we perform shape completion of sparse, partial point clouds, and, in this context, we compare and analyze existing related methods. In \cref{sec:modalities}, we demonstrate different applications of our method by generating samples from real scanned, noisy point clouds, and 2D images. We end with an ablation study validating the design choices in \cref{sec:ablation}.

%alternatively, maybe list training hardware and times and move the rest to supplement
\paragraph{Implementation} We train our method as follows. First, we train our joint SDF-VAE model on full point clouds and corresponding query points and ground truth signed distance values. We combine the architecture of GenSDF~\cite{gensdf} with a VAE~\cite{vae} consisting of a 5-layer encoder and 5-layer decoder. The PointNet in GenSDF outputs three 2D plane features from the point cloud, which we concatenate and pass as input to the VAE. The bottleneck of the VAE ($z$ in \cref{fig:training}) is a 1D latent embedding of an SDF, and we save them after training is complete. Next, we use the latent vectors as training data for the diffusion model. We follow the architecture of DALLE-2~\cite{dalle2, github-dalle2}. There are six blocks each consisting of a self-attention layer and a fully-connected layer. For conditional training, we introduce custom encoders for different inputs. We use a PointNet~\cite{pointnet} for point clouds and ResNet 18~\cite{resnet} for 2D images. We also add another cross-attention layer to each block of the diffusion model that learns key and value pairs from extracted shape features. Finally, we fine-tune both modules end-to-end by connecting them as illustrated in \cref{fig:training}. We provide a full architecture description and training details in the supplement. 

\paragraph{Datasets} 
For unconditional generation and partial point cloud completion, we train and evaluate using Acronym~\cite{acronym}. Acronym is a processed subset of the popular ShapeNet~\cite{shapenet} dataset and contains watertight, synthetic 3D meshes across 262 shape categories. We use three training splits. The first two are single categories: Chair and Couch. The third split uses all classes that have at least 20 objects, providing us with 106 classes in total. From each of them, we take at most 50 objects to prevent the model from overfitting to larger categories. For single-view reconstruction, we use the Airplane and Couch categories from ShapeNet~\cite{shapenet} and their rendered 2D images as conditioning input. For real-scanned point clouds, we use YCB~\cite{ycb}, a collection of point clouds acquired from multi-view RGBD captures.
% For unconditional generation, we sample from a random standard Gaussian distribution $\mathcal{N}(0,\textbf{I})$. For conditional generation, we guide the random Gaussian with point clouds from this training set as well as from unseen objects and categories. 

\paragraph{Evaluation}
For unconditional generation, we follow Yang \etal~\cite{pointflow} and use minimum matching distance (MMD), coverage (COV), and 1-nearest neighbor accuracy (1-NNA). MMD measures quality, COV measures diversity, and 1-NNA uses a classifier to measure the similarity of the reference and generated distributions, where $50\%$ accuracy means the generated set is indistinguishable from the reference set. We generate the same number of samples as the reference set. For conditional generation (shape completion), we follow Wu \etal~\cite{cgan} and evaluate MMD, total mutual difference (TMD), and unidirectional Hausdorff distance (UHD). TMD measures diversity and UHD measures fidelity to the input partial point cloud. We generate 10 samples for every input partial point cloud in the reference set. For all metrics except UHD, we use Chamfer Distance (CD)~\cite{pointnet} as the distance measure. We extract 2,048 points from each sample to calculate these metrics. Additionally, since generated results are random, we run evaluation 5 times and report the best set of metrics. We provide detailed formulations of all metrics in the supplement. For visualization, we run Marching Cubes~\cite{marching-cubes} and render the resulting meshes.

\begin{figure}[t]
    \centering 
    \includegraphics[width=\linewidth]{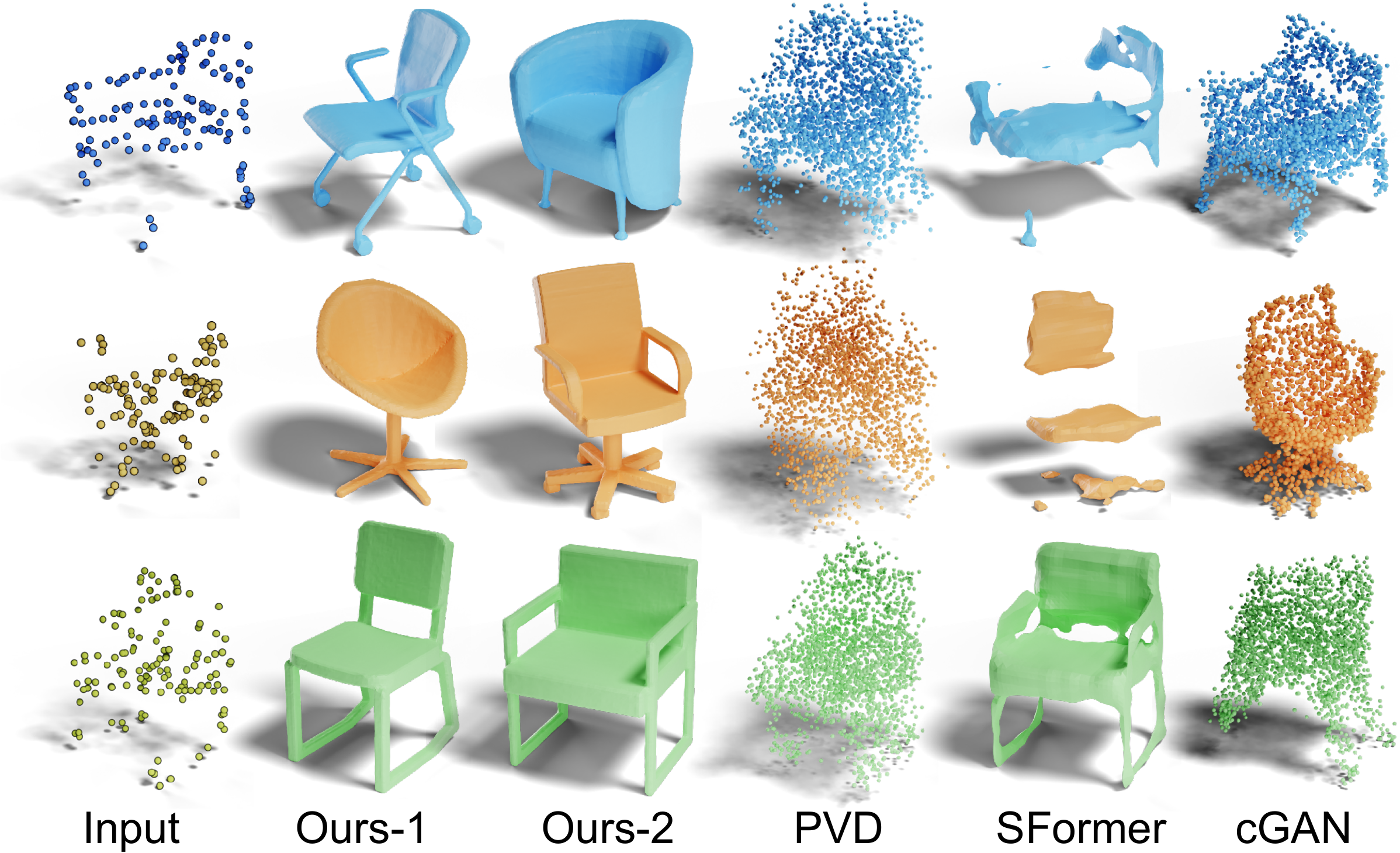}
    \caption{Shape completion results from sparse, partial point clouds. Reconstructions from the proposed method represent details such as the legs of the chair, whether they are separated (top), branched out (middle), or connected (bottom).}
    \label{fig:cond-compare}
\end{figure}

% \begin{figure}[t]
%     \centering 
%     \includegraphics[width=\linewidth]{figs/results/cond-diversity.pdf}
%     \caption{Our shape completion results are more diverse than those of existing works (see \cref{tab:conditional}). Given an input point cloud (left-most column) that indicates a ``long" couch (top row) or ``bowl shape" (bottom row), our model produces distinct samples.}
%     \label{fig:cond-diversity}
% \end{figure}

\subsection{Unconditional Generation}
\label{sec:unconditional}
We train unconditional models on the three data splits mentioned above: Chair, Couch, and Multi-class. We compare to ShapeGAN~\cite{shapegan}, which generates SDFs, and PVD~\cite{pvd} and DPM3D~\cite{pc-diffusion}, which both diffuse point clouds. Our method outperforms baselines in all metrics as reported in \cref{tab:unconditional}. Furthermore, our method has substantially higher diversity measured by coverage (COV), surpassing the second-best result by roughly $10\%$ in all experiments. This is due to our regularized latent space, which allows the model to learn and interpolate from a continuous distribution. Our visualizations (\cref{fig:teaser} and \cref{fig:uncond-results}) validate that our model generates clean 3D surfaces. We also calculate the average distance between the generations and each object in the reference set to confirm that our model is capable of producing diverse and unique shapes.

\subsection{Conditional Generation for Shape Completion}
\label{sec:conditional}

Next, we assess the proposed method for shape completions of sparse, partial point clouds. During both training and testing, we randomly sample 128 points from a full point cloud, then crop $50\%$ of points. We randomly select a viewpoint and remove the 64 furthest points from the viewpoint to obtain a partial point cloud, following Yu \etal~\cite{pointr}. We perform this cropping during training online in each iteration. We report quantitative results in \cref{tab:conditional} and show visualizations in \cref{fig:cond-compare}. Previous works perform well on dense partial point clouds and we show results in our supplement. However, completing a sparse and partial point cloud is challenging. Methods such as ShapeFormer (SFormer)~\cite{shapeformer} and AutoSDF~\cite{autosdf} fail under this setting because they quantize shapes into patches. Sparse point clouds mean there are very few patches to extract information from. PVD~\cite{pvd} produces noisy samples because they operate on discrete points and cannot interpolate smoothly from a learned prior distribution. cGAN~\cite{cgan} learns a regularized latent space so its generations are relatively clean and complete but are less diverse because its priors are less expressive due to the ``prior hole problem" of VAEs, which diffusion models solve~\cite{lion, prior-hole1,prior-hole2, prior-hole3}. 

\begin{figure}[t]
    \centering 
    \includegraphics[width=\linewidth]{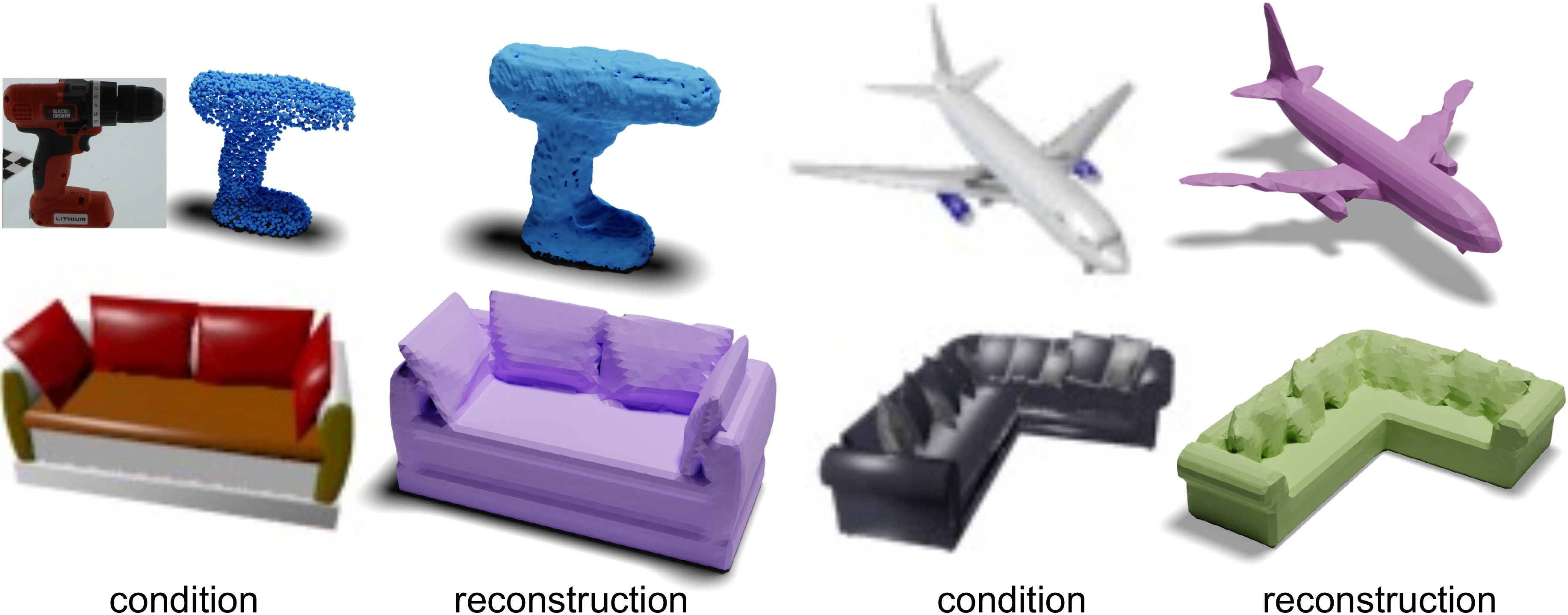}
    \caption{Reconstructing scanned point clouds and single images. Our method captures details of conditioned geometry, such as the curves of the drill, engines of the plane, and pillows on the couch.}
    \label{fig:modalities}
    %\vspace{-0.5em}
\end{figure}

Our method outperforms all baselines in MMD (quality) and TMD (diversity) but not UHD. 
% However, we find that UHD may not be a good indicator of fidelity. 
The UHD metric measures fidelity by finding the largest distance between any partial input point and its nearest neighbor to the completed shape, so outliers determine the UHD value. We note that PVD~\cite{pvd} performs well under the UHD metric but generations are noisy and less realistic.
%Additionally, we only sample 2048 points because the output points of some baselines are limited, but this is not an accurate approximation of the surface.
Our visualizations in \cref{fig:cond-compare} show that our completions match the input well. Given indication of the style of the legs of a chair, our method produces plausible shapes accordingly. 

% In future work, we could enforce the relation between the latent vectors of partial shapes and those of complete shapes. Then, by fixing the latent vector of the partial shape, we could enforce its presence in the output. 

\subsection{Other Modalities for Conditioning}
\label{sec:modalities}
In our method, we formulate shape completion, single-view reconstruction, and reconstruction of real-scanned point clouds as a unified task. Essentially, we are learning a distribution over 3D shapes that we can sample from, given a condition. In \cref{fig:teaser} and \cref{fig:modalities}, we show two additional modalities: real scanned, noisy point clouds and 2D images. A sample image for extracting the scanned point cloud is shown for reference, but was not used during training. For the former, we train from scratch using the YCB~\cite{ycb} dataset, a collection of point clouds acquired from multi-view RGBD captures. 
%The fused multi-view point clouds resemble collection of in-the-wild data for tasks such as robotic grasping. 
The point clouds are noisy and incomplete (e.g., the bottom of each object is on a table and not captured). For conditioning on 2D images, we use a pretrained ResNet 18~\cite{resnet} as our encoder $\Upsilon$. 
%Our model remains capable of reconstructing plausible outputs for both tasks, shown in \cref{fig:modalities}. We provide training details in our supplement. 

\begin{figure}[t]
    \centering 
    \includegraphics[width=0.9\linewidth]{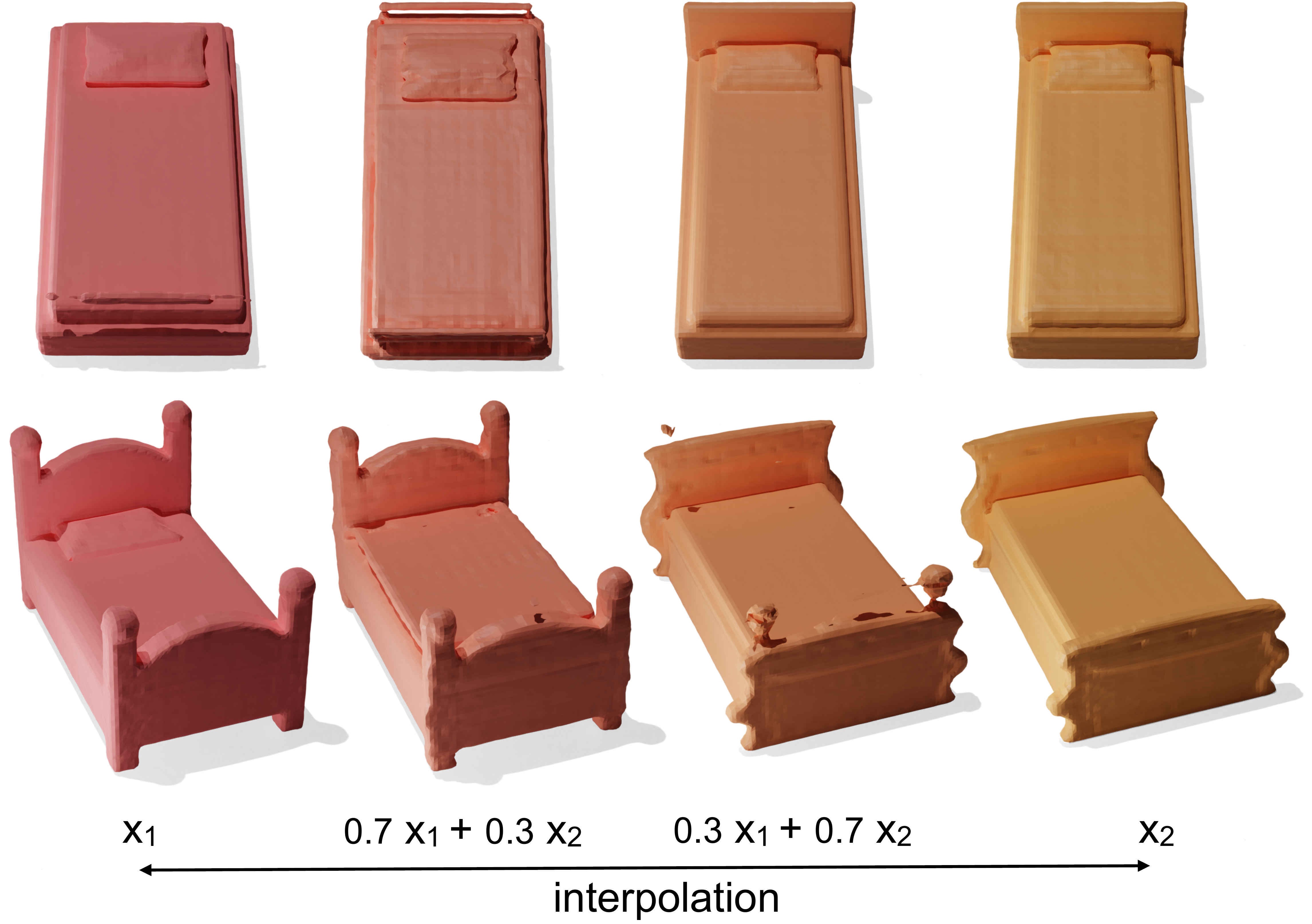}
    \caption{Interpolation of generations $x_1, x_2$ by performing a linear combination of their respective latent vectors. The resulting vectors are decoded and meshed.}
    \label{fig:interpolation}
    \vspace{-1em}
\end{figure}

\subsection{Scalability and Interpolation}
\paragraph{Large Training Datasets} We investigate whether our VAE poses a bottleneck to learning large datasets. The single-category experiments (Chair, Couch) are performed on $558$ and $366$ meshes, respectively. Our multi-class split contains $4230$ meshes. As such, without adjusting the architecture or number of parameters, our method scales without degradation in quality or creating artifacts. To further validate scalability, we experiment with $90\%$ of the entire Acronym dataset, $7148$ meshes in total. The average Chamfer Distance (CD) of the SDF reconstructions of all $7148$ meshes is $0.92 \x 10^{-3}$, and $0.87 \x 10^{-3}$ when reconstructing only the Couch category. This value is \emph{lower} than training on the single Couch category, where the reconstruction CD is $1.04 \x 10^{-3}$. This validates that our approach scales gracefully, learns better when we introduce more training data, and generalizes to out-of-distribution shapes as many categories have very few data (just 1-10 meshes) compared to the larger classes ($300$-$500$ meshes). See visualizations in the supplement.
%We provide additional discussion in the supplement. 
%couch average/median: $0.00104/0.00607$
%couch average/median from the larger dataset: 0.000874, 0.000835
%entire larger dataset: 0.000924, 0.000778
\vspace{-1em}
%ADD BETTER RESULTS AFTER TRAINING LARGE DATASET
\paragraph{Latent Interpolation} In \cref{fig:interpolation}, we interpolate between generated samples by performing a linear combination of their respective latent vectors. Gradual shift in shape features, such as the headboards, validate that our learned latent space is continuous and the latents control semantic geometry. Thus, during generation, our model randomly samples from the latent space to form novel shapes. We provide further discussion in the supplement.

\subsection{Ablation Experiments}
\label{sec:ablation}
We analyze design choices that affect conditional generations. In \cref{tab:ablation}, we report shape completion metrics for the Couch training split. We also define a \textit{consistency (CONS)} metric for measuring fidelity, following our observation in \cref{sec:conditional} that UHD does not correlate closely with visual results. For CONS, we evaluate all points in the input partial point cloud using the generated SDF and take the average of the predicted signed distance values. If the points are to be present on the reconstructed surface, then by definition, the values of each point are close to 0. This metric also allows us to filter generations before running marching cubes, which reduces inference time while maintaining high-quality samples. For filtering, we sample the maximum number of meshes that can fit into one sampling batch (30 in our case), and keep the 10 meshes with the lowest consistency scores. 
Note that the CONS filter is only used for conditional generation, and not during training.
% We do not use the CONS filter during training, only for conditional generations. 

\textit{No end-to-end} refers to skipping the fine-tuning stage explained in \cref{sec:endtoend}. Generations are clean and realistic but lack diversity and complexity. This confirms our end-to-end training scheme improves generalization and introduces geometrical constraints. Next, $\omega$ refers to the ratio of guidance strength during generation: the final output is a linear combination of two generations with and without guidance: $(\omega+1)z_{c}-(\omega)z_u$, where $z_{c}$ is guided and $z_{u}$ is unconditional~\cite{classifier-free-conditional-dropout}. In our experiments, we guide generations every iteration (i.e., $\omega$ = $0$) but depending on the use case, one can adjust this hyperparameter to determine the tradeoff between diversity and fidelity. See supplement for details. Then, we experiment with \textit{Concatenation} instead of cross-attention for conditioning, following \cite{dalle2}. We diffuse the concatenation of the conditioned feature and noisy vector and rely on self-attention. We find both conditioning mechanisms lead to similar outputs, but concatenation increases dimensions and memory consumption significantly. Finally, we report quantitative values without filtering in \textit{no filt}. We show that our CONS filter does not inflate MMD (quality) and TMD (diversity). We note that our model prioritizes generation quality and diversity, at the cost of the UHD metric. In future work, we plan to enforce the relationship between latent vectors of partial shapes and those of complete shapes to improve consistency.

% On the right side of \cref{tab:ablation}, we validate our modulation module, which is the cornerstone of our SDF representation. As mentioned in \cref{sec:related_work}, Dupont \etal~\cite{functa} use SIREN and meta-learning. On single categories, this method performs adequately, but completely fails to represent detailed shapes when we increase the number of categories. We also experiment with auto-decoders and arrive at a similar result. These two methods rely on linear mappings between discrete latent codes and a base SDF network, and the distribution of shapes they can represent is limited. We show visualizations in the supplement. The proposed method learns a regularized latent space and a shape prior, which represents over 100 categories with fine details. In \cref{sec:method}, we mention that we use a KL-loss to enforce the target latent distribution to be a zero-mean Gaussian with standard deviation 0.25. In \textit{Ours ($\sigma=1.0$)} and \textit{Ours ($\sigma=0.5$)} we experiment with standard deviations 1.0 and 0.5, respectively. Increasing $\sigma$ can lead to an overly spread-out distribution that cannot encode shapes well, although we speculate that on datasets substantially larger than the ones we use, $\sigma$ could be increased. Apart from representation, our current choice of 0.25 allows the diffusion model to learn from a more compact distribution, which improved convergence speeds and training stability. 

\begin{table}[t]
    \caption{Ablation study based on conditional generation using the Couch split. $\uparrow$ means higher is better and $\downarrow$ means lower is better. All values are scaled up by $10^2$.}
    \centering
    \resizebox{0.9\linewidth}{!} {%
      \begin{tabular}{lccc}
    \toprule
     & MMD $(\downarrow)$ & TMD $(\uparrow)$ & CONS $(\downarrow)$ \\
    \midrule 
    No end-to-end & $0.096$& $8.292$& $5.346$   \\
    \midrule
    $\omega=4$ & $0.044$& $7.251$& $\textbf{0.822}$  \\
    $\omega=1$& $0.049$& $11.46$& $1.594$   \\
  \midrule 
  Concatenation  & $0.049$& $12.22$& $2.068$   \\
  \midrule 
    Ours & $\textbf{0.041}$& $13.53$& $1.967$  \\
    Ours (no filt) & $\textbf{0.041}$& $\textbf{17.06}$& $3.545$  \\
    \bottomrule
    \end{tabular} %
    }
    \vspace{-1em}
    \label{tab:ablation}
\end{table}

\section{Conclusion}
We devise a probabilistic diffusion model that generates diverse shapes from a distribution of learned SDFs. We assess the proposed method for shape generation and completion of various input modalities. %Given the method's potential downstream applications and generation quality, we hope this work brings us closer to democratizing creation of 3D assets through AI. 
To further improve the method, we could speed up inference time of the diffusion model with techniques such as DDIM sampling~\cite{ddim}, and we could enforce the relationship between the latents of partial shapes and those of complete shapes to improve interpretability and fidelity. There are also many avenues for exciting future work. Besides exploring other conditioning approaches, e.g., text-to-shape, we could learn appearance for generating realistic assets. We would also be interested in expanding Diffusion-SDF to full scene synthesis.

{\small
\bibliographystyle{ieee_fullname}
\bibliography{bib}
}
% \clearpage
% \foreach \x in {1,...,12}{
% \includepdf[pages={\x}]{supp_pdf}
% }
% \includepdf[pages={-}]{supp_pdf}
% \includepdf[pages={2}]{supp_pdf}
% \foreach \x in {1,...,6}{
% \begin{figure}
%   \includegraphics[page=\x]{supp_pdf}
%   \clearpage
% \end{figure}
% }

\end{document}